\documentclass[conference]{IEEEtran}
\IEEEoverridecommandlockouts

\usepackage{amsmath,amssymb,amsfonts}
\usepackage{graphicx}
\usepackage{textcomp}
\usepackage{xcolor}
\usepackage{booktabs}
\usepackage{tabularx}
\usepackage{url}
\usepackage{tikz}
\usetikzlibrary{arrows.meta,positioning,calc}

\usepackage{hyperref}
\hypersetup{
  hidelinks,
}

\graphicspath{{figures/}}

\def\BibTeX{{\rm B\kern-.05em{\sc i\kern-.025em b}\kern-.08em
    T\kern-.1667em\lower.7ex\hbox{E}\kern-.125emX}}

\begin{document}

\title{W4A4 Quantization for Inference on Wan2.2-I2V-A14B\\
{\large ICME 2026 Low-Bit-width Large-Model Quantization Challenge\\
(Sub-Challenge~1: HiF4 / MXFP4)}}

\author{\IEEEauthorblockN{Yidong Chen\IEEEauthorrefmark{1}\thanks{Code: \url{https://github.com/shch-y/icme}.},
Chengyu Shi\IEEEauthorrefmark{1},
Jiahao Liu\IEEEauthorrefmark{1}}
\IEEEauthorblockA{\IEEEauthorrefmark{1}Tsinghua University\\
ICME 2026 Low-Bit-width Large-Model Quantization Challenge team submission}}

\maketitle

\begin{abstract}
We summarize our submission to Sub-Challenge~1: W4A4 Quantization for Inference (HiF4 / MXFP4) of the ICME~2026 Low-Bit-width Large-Model Quantization Challenge.
The sub-challenge targets 4-bit weight and 4-bit activation (W4A4) inference on Wan-AI/Wan2.2-I2V-A14B under HiF4 or MXFP4 numerical formats.
We adapt two complementary ideas from LLM quantization---MixQ-style mixed precision for sparse activation outliers and SmoothQuant-style per-channel smoothing---together with block-wise HiF4 packing for Wan2.2 feed-forward (FFN) linear layers.
Calibration on representative OpenS2V-5M batches identifies heavy-tailed activation channels; smoothing rebalances dynamic range before W4A4 rounding; and a dual-branch GEMM preserves outlier columns in higher precision while the bulk of channels use strict W4A4.
On official VBench I2V metrics, our pipeline stays within 2--3.5\% of FP16 on most quality axes and improves motion smoothness, outperforming a native HiFloat4 baseline that degrades roughly 5\% relative to FP16 across all reported scores.
\end{abstract}


\section{Introduction}

Image-to-video (I2V) diffusion models such as Wan2.2-I2V-A14B~\cite{wan22} deliver strong generative quality but impose substantial memory and compute costs at \textbf{inference} time.
Sub-Challenge~1 of the ICME~2026 quantization grand challenge therefore focuses on \textbf{W4A4} inference: quantizing both weights and activations of linear layers to 4-bit HiF4/MXFP4-compatible formats while preserving perceptual quality measured by \textbf{OpenS2V-5M} prompts and \textbf{VBench} I2V metrics.
Organizers allow a limited number of Transformer blocks to remain in high precision (at most five layers for MXFP4 and two for HiF4); we target the main Wan2.2 track and recommend generation at $720{\times}1280{\times}61$ frames.

Direct W4A4 rounding on video diffusion transformers is difficult for the same reason it is difficult in large language models (LLMs): activation tensors contain \emph{heavy-tailed outliers}---a tiny fraction of channels carries magnitudes far larger than the bulk distribution, which inflates per-tensor scales and destroys 4-bit signal-to-noise ratio (SNR), especially inside FFN projections.
Prior LLM systems address this along two largely orthogonal axes.
\textbf{SmoothQuant}~\cite{xiao2023smoothquant} \emph{migrates} quantization difficulty from activations to weights via an exact per-channel fold, making W8A8 (and, with care, W4A4) feasible without retraining.
\textbf{MixQ}~\cite{mixq} \emph{splits} computation into a high-throughput low-bit bulk path and a narrow high-precision outlier path, using locality-based prediction in LLMs to avoid expensive runtime outlier detection.

\begin{figure}[t]
  \centering
  \includegraphics[width=\linewidth]{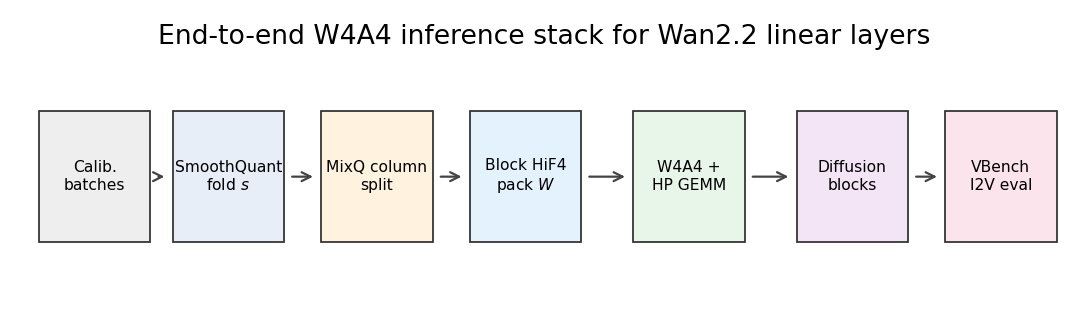}
  \caption{End-to-end W4A4 inference pipeline: offline calibration, SmoothQuant fold, MixQ column split, block-packed weights, and dual-branch GEMM before VBench evaluation.}
  \label{fig:pipeline}
\end{figure}

Our submission transfers these insights to Wan2.2 I2V under a fixed inference contract: we apply SmoothQuant-style folding first, then MixQ-style column splitting on FFN linear maps, and finally block-wise HiF4 packing for weights.
Fig.~\ref{fig:pipeline} summarizes the end-to-end stack evaluated in this report.

\section{Background: Outliers, SmoothQuant, and MixQ}

\subsection{Why activations dominate W4A4 error}

\begin{figure}[t]
  \centering
  \includegraphics[width=\linewidth]{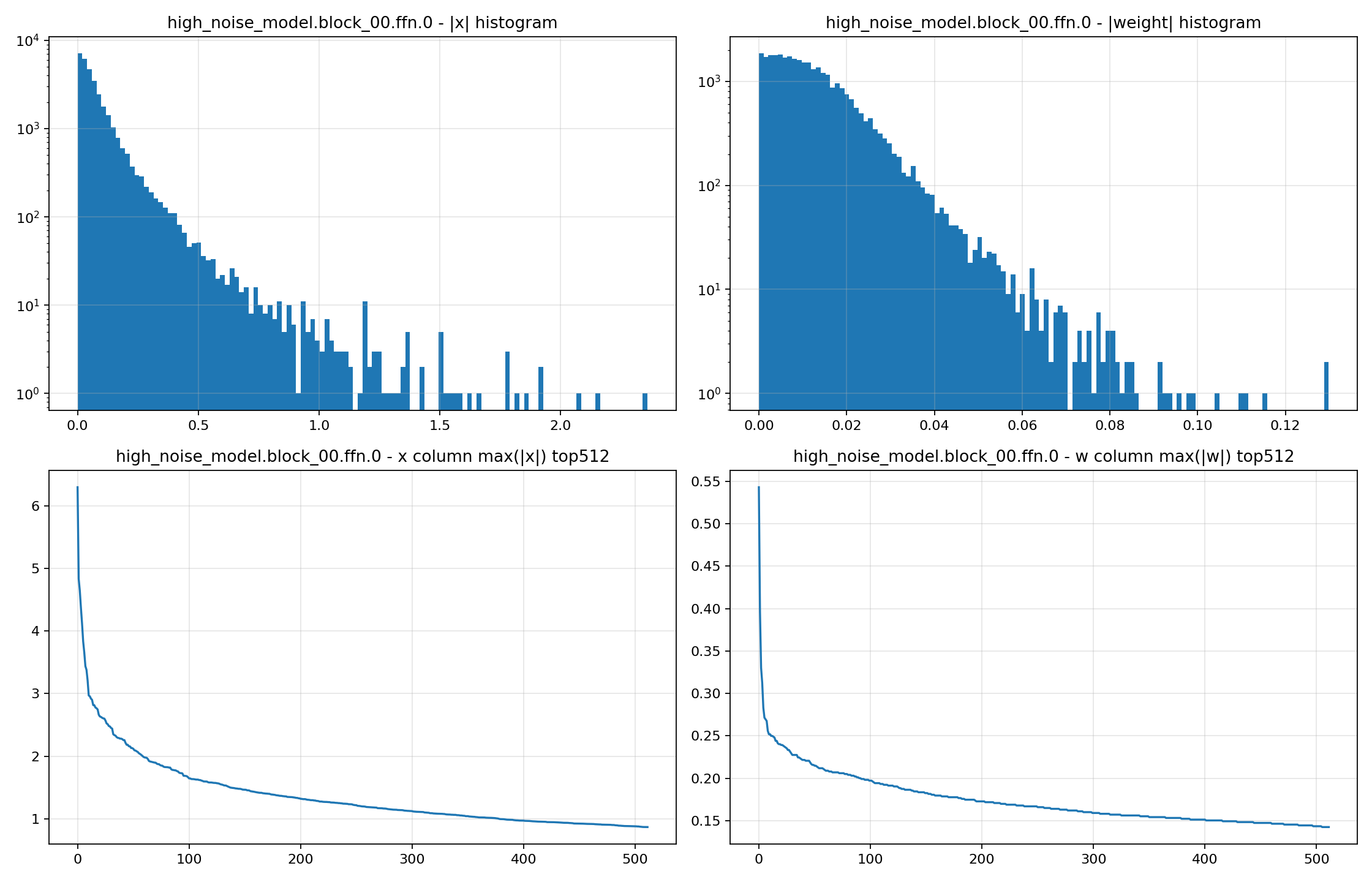}
  \caption{Wan2.2 FFN activation/weight statistics (\texttt{block\_00.ffn.0}): heavy-tailed $|x|$ and sparse high-magnitude columns motivate mixed-precision and smoothing before W4A4.}
  \label{fig:wan-ffn}
\end{figure}

For a linear layer $Y=XW$, weight tensors in Wan2.2 are comparatively smooth, whereas FFN activations exhibit sharp per-channel maxima.
Fig.~\ref{fig:wan-ffn} shows calibration statistics from \texttt{high\_noise\_model.block\_00.ffn.0}: activation column maxima reach above $6.0$ while weight column maxima stay below $0.6$, and the $|x|$ histogram has a long tail beyond $|x|{>}1.5$.
Under per-tensor or coarse per-channel W4A4, such tails set the quantizer step size and compress the majority of small-valued channels into a few bins.

\subsection{SmoothQuant recap}

\begin{figure}[t]
  \centering
  \includegraphics[width=\linewidth]{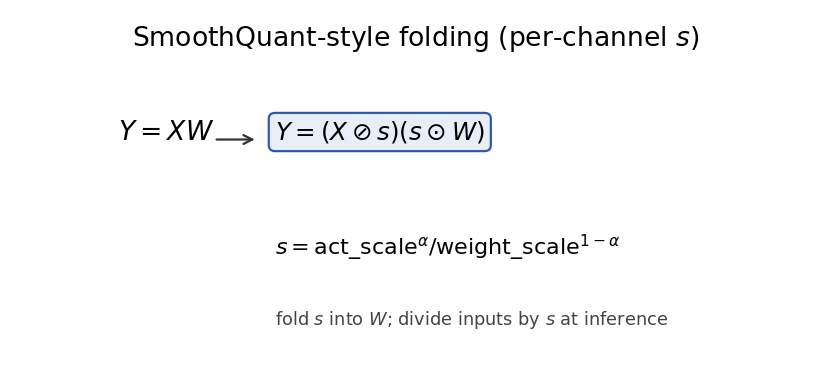}
  \caption{SmoothQuant folding preserves $Y=XW$ in floating point while rebalancing outliers between $X$ and $W$.}
  \label{fig:smoothquant}
\end{figure}

\begin{figure}[t]
  \centering
  \includegraphics[width=\linewidth]{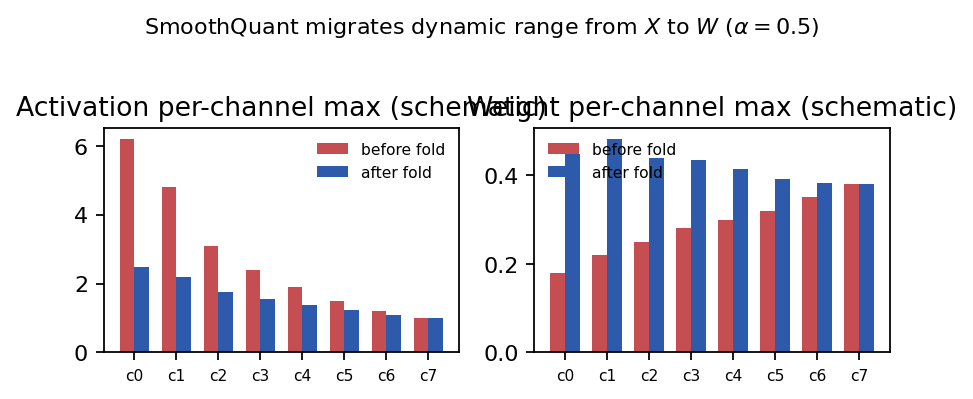}
  \caption{Schematic per-channel maxima before/after SmoothQuant fold ($\alpha{=}0.5$): activation tails shrink while corresponding weight columns grow.}
  \label{fig:sq-migrate}
\end{figure}

SmoothQuant~\cite{xiao2023smoothquant} introduces a positive per-channel scale $s$ such that
\begin{equation}
Y = \bigl(X \oslash s\bigr)\,\bigl(s \odot W\bigr),
\label{eq:sq}
\end{equation}
where $\oslash$ and $\odot$ apply along input channels.
During calibration, activation and weight magnitudes are collected and combined as
\begin{equation}
s_j = \frac{\text{act\_scale}_j^{\alpha}}{\text{weight\_scale}_j^{\,1-\alpha}},
\qquad \alpha \in (0,1).
\label{eq:smooth}
\end{equation}
Larger $\alpha$ pushes more dynamic range into $W$; smaller $\alpha$ leaves more in $X$.
At inference, $s$ is absorbed into folded weights while activations are divided by $s$ on the fly, reducing the activation range \emph{exactly} before HiF4/MXFP4 rounding.
Fig.~\ref{fig:smoothquant} and Fig.~\ref{fig:sq-migrate} illustrate the fold and the intended per-channel rebalancing.

\subsection{MixQ-style mixed precision recap}

Mixed-precision quantization~\cite{mixq} keeps a small outlier set in high precision and quantizes the remainder to INT4/INT8 or floating 4-bit formats.
For activation $A$ with outlier channel set $\mathcal{O}$, a generic mixed GEMM can be written as
\begin{equation}
\begin{split}
C_{i,j} \approx\;& S^{A}_{i} S^{W}_{j} \!\!\sum_{k \notin \mathcal{O}}\! (A_{\mathrm{q}})_{i,k} (W_{\mathrm{q}})_{j,k} \\
& + \sum_{k \in \mathcal{O}} (A_{\mathrm{hp}})_{i,k} (W_{\mathrm{hp}})_{j,k},
\end{split}
\label{eq:mix-gemm}
\end{equation}
where $A_{\mathrm{q}},W_{\mathrm{q}}$ are low-bit tensors with per-group scales $S^{A},S^{W}$.
MixQ's original LLM system further exploits \emph{token-level locality}: outlier channels are predictable across consecutive decode steps, enabling online prediction and a \emph{quantization-ahead-of-detection} fast path that avoids atomic outlier scans on most tokens~\cite{mixq}.

Video diffusion inference differs: tokens correspond to spatio-temporal tokens across denoising steps, and the challenge mandates a fixed HiF4/MXFP4 W4A4 kernel rather than a custom INT8/FP16 outlier micro-kernel.
We therefore retain the \textbf{outlier split} and dual-branch structure of MixQ, but calibrate a \textbf{static} top-$k$ column set per FFN layer from OpenS2V-5M batches, analogous to ahead-of-time outlier selection yet guided by the same $\max_t|x_{t,j}|$ score used in MixQ's channel view.

\begin{figure}[t]
  \centering
  \includegraphics[width=\linewidth]{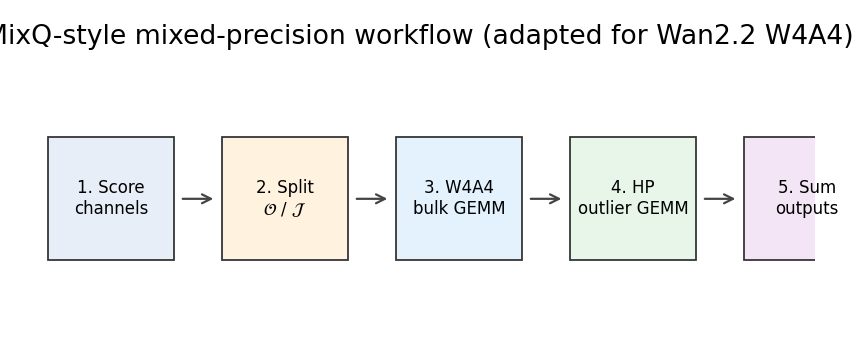}
  \caption{MixQ-style workflow adapted to Wan2.2 W4A4: score channels, split $\mathcal{O}/\mathcal{J}$, run bulk W4A4 GEMM and a narrow high-precision branch, then sum.}
  \label{fig:mixq-flow}
\end{figure}

\section{Method}

\subsection{Per-layer calibration protocol}

We collect activation and weight statistics from a small set of OpenS2V-5M calibration prompts at the organizer-recommended resolution.
For each targeted FFN linear map we record per-input-channel activation maxima, weight column ranges, and optional block-wise extrema for HiF4 packing.
All offline steps run once; deployed inference executes only folded weights, fixed column masks $\mathcal{O}$, and W4A4/HiF4 kernels.

\subsection{SmoothQuant-style folding for Wan2.2 FFN}

\begin{figure}[t]
  \centering
  \includegraphics[width=\linewidth]{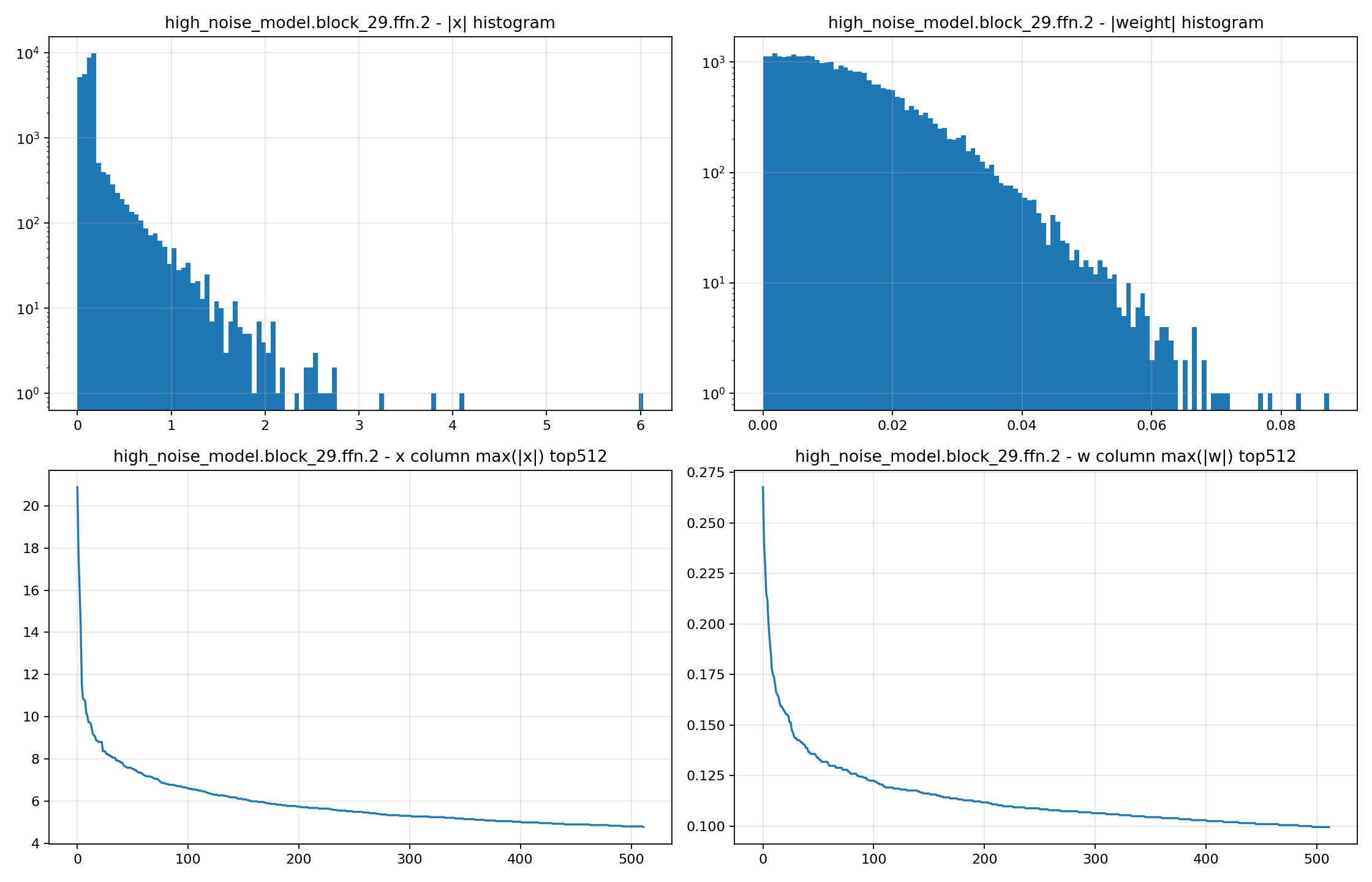}
  \caption{FFN statistics at \texttt{block\_29.ffn.2}: outlier structure persists in deep layers, supporting a shared SmoothQuant+MixQ recipe.}
  \label{fig:wan-ffn-deep}
\end{figure}

We apply Eq.~\eqref{eq:smooth} layer-wise with $\alpha$ tuned on calibration clips (typically $0.5$--$0.7$ for FFN up-projections where activation tails are strongest).
Folded weights are re-packed into block HiF4 tiles; runtime activations are divided by the same $s$ immediately before 4-bit activation quantize.
This step is applied \emph{before} MixQ splitting so that both bulk and outlier branches see reduced activation tails.
Deeper blocks exhibit similar but attenuated tails (Fig.~\ref{fig:wan-ffn-deep}); the same recipe is applied uniformly unless a layer is kept in FP16 under organizer rules.

\subsection{MixQ-style column split and dual-branch GEMM}

\begin{figure}[t]
  \centering
  \includegraphics[width=0.95\linewidth]{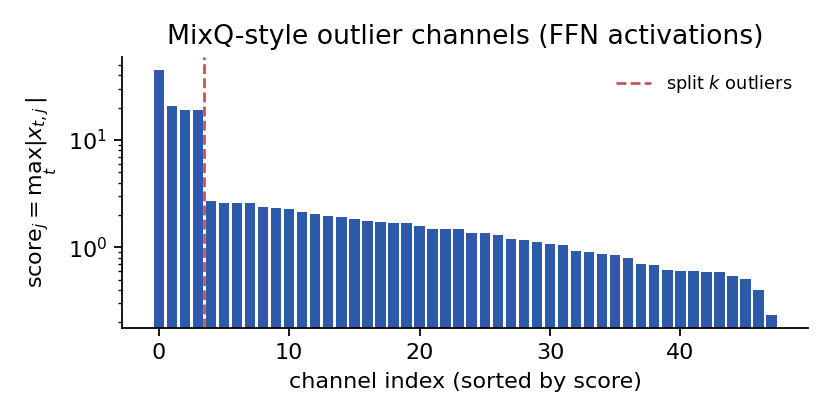}
  \caption{Sorted per-channel $\text{score}_j$ on an FFN projection: a few dominant columns justify a MixQ-style split.}
  \label{fig:mixq-scores}
\end{figure}

For $X\!\in\!\mathbb{R}^{T\times d_{\mathrm{in}}}$ we rank input channels by
\begin{equation}
\text{score}_j = \max_t \bigl| x_{t,j} \bigr|
\label{eq:score}
\end{equation}
and form $\mathcal{O}$ as the top-$k$ columns (Fig.~\ref{fig:mixq-scores}).
Let $\mathcal{J}=\{1,\ldots,d_{\mathrm{in}}\}\setminus\mathcal{O}$.
The layer output is
\begin{equation}
Y = X_{:,\mathcal{J}} W^{\mathrm{(q)}}_{\mathcal{J},:} + X_{:,\mathcal{O}} W^{\mathrm{(hp)}}_{\mathcal{O},:},
\label{eq:dual}
\end{equation}
where $(\cdot)^{\mathrm{(q)}}$ denotes HiF4/MXFP4 W4A4 and $(\cdot)^{\mathrm{(hp)}}$ is FP16/BF16.
$k$ is chosen to respect the challenge's high-precision budget: a handful of columns per FFN map adds negligible FLOPs but prevents the largest tails from collapsing under 4-bit activations.
Conceptually, Eq.~\eqref{eq:dual} is the Wan2.2 specialization of Eq.~\eqref{eq:mix-gemm} with a static $\mathcal{O}$.

\subsection{Block-wise HiF4 weight packing}

\begin{figure}[t]
  \centering
  \includegraphics[width=\linewidth]{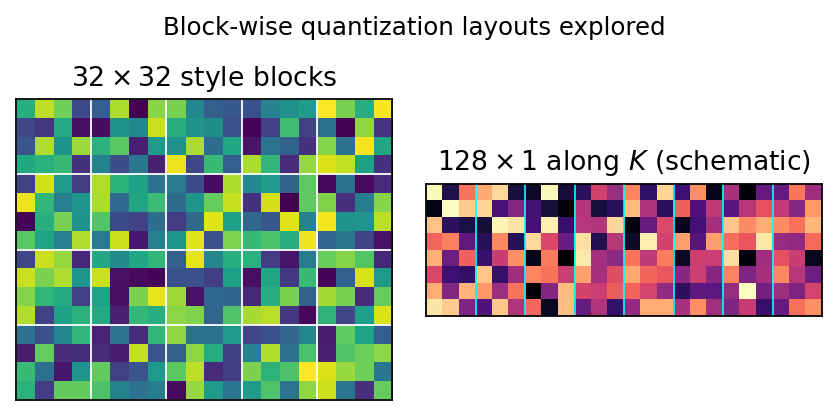}
  \caption{Block layouts explored for HiF4/MXFP4 weight packing after SmoothQuant fold.}
  \label{fig:block}
\end{figure}

After folding and optional outlier column extraction on $W$, remaining weights are packed with block-wise scales~\cite{krishnamoorthi2018}.
We evaluate $32{\times}32$ tiles and $128{\times}1$ $K$-major stripes (Fig.~\ref{fig:block}); the former improves SNR on outlier-heavy tiles, while the latter matches GEMM memory order.
Per-block exponents are stored alongside 4-bit payloads as required by HiF4/MXFP4 simulation kernels.

\subsection{Design rationale: why combine SmoothQuant and MixQ?}

SmoothQuant and MixQ attack different failure modes.
SmoothQuant is \emph{global} and \emph{linear}: it rebalances every channel but cannot fully eliminate the heaviest tails that remain after fold (Fig.~\ref{fig:wan-ffn}, bottom-left).
MixQ is \emph{selective}: it spends high-precision budget only where tails survive smoothing.
In LLM serving, MixQ additionally removes detection overhead via online prediction; in our challenge setting, static $\mathcal{O}$ keeps the inference graph simple while still capturing the same outlier energy that MixQ was designed to preserve.
Applying smoothing before splitting also shrinks the high-precision branch: fewer columns exceed the W4A4 representable range after fold, so $k$ can stay small.
In short, SmoothQuant stabilizes the W4A4 bulk path, MixQ protects residual high-magnitude columns with small $|\mathcal{O}|$, and block-wise packing provides finer scales for local weight outliers. The full stack is used for the final OpenS2V-5M submission.

\subsection{Implementation notes}

All three stages are implemented as offline graph transforms plus inference-time hooks in the Wan2.2 high-noise and low-noise transformer stacks.
Calibration reuses the same OpenS2V-5M prompts as generation; statistics are accumulated only on FFN \texttt{nn.Linear} inputs after SiLU/GELU gates.
SmoothQuant scales are folded into master weights before HiF4 packing, and column masks $\mathcal{O}$ are saved as sparse index lists per layer.
At inference, activations are divided in-place by $s$, quantized to HiF4, and routed through either the W4A4 bulk path or a narrow FP16 outlier path.
This mirrors MixQ~\cite{mixq}, but trades online token prediction for deterministic masks that are easy to validate under challenge rules.

\section{Evaluation and Results}

\subsection{Protocol}

We follow the challenge OpenS2V-5M evaluation split and report VBench I2V metrics: aesthetic quality, I2V subject alignment, imaging quality, motion smoothness, and subject consistency.
VBench-I2V does not provide T2V \emph{overall consistency}; we omit it.
Table~\ref{tab:vbench} lists FP16, a native HiFloat4 W4A4 reference, and our full pipeline; parenthetical percentages are relative change vs.\ FP16.

\begin{table}[t]
  \centering
  \caption{VBench I2V metrics vs.\ FP16. Parentheses: relative change (\%). Native HiFloat4 $\sim$5\% below FP16 on each metric.}
  \label{tab:vbench}
  \scriptsize
  \setlength{\tabcolsep}{2.5pt}
  \begin{tabular}{lccccc}
    \toprule
    Model & Aesthetic & I2V subj.\ & Imaging & Motion & Subj.\ cons. \\
    \midrule
    FP16 baseline
      & \textbf{0.5445} & \textbf{0.9626} & \textbf{0.7086} & 0.9730 & \textbf{0.9199} \\
    Native HiFloat4
      & \shortstack{0.5173\\{\tiny $-$5.0\%}}
      & \shortstack{0.9145\\{\tiny $-$5.0\%}}
      & \shortstack{0.6732\\{\tiny $-$5.0\%}}
      & \shortstack{0.9243\\{\tiny $-$5.0\%}}
      & \shortstack{0.8739\\{\tiny $-$5.0\%}} \\
    Ours (W4A4)
      & \shortstack{0.5274\\{\tiny $-$3.1\%}}
      & \shortstack{0.9375\\{\tiny $-$2.6\%}}
      & \shortstack{0.6936\\{\tiny $-$2.1\%}}
      & \shortstack{\textbf{0.9769}\\{\tiny $+$0.4\%}}
      & \shortstack{0.8875\\{\tiny $-$3.5\%}} \\
    \bottomrule
  \end{tabular}
\end{table}

\subsection{Discussion}

Native HiFloat4 W4A4 without outlier handling or smoothing loses $\sim$5\% on every axis relative to FP16 (Table~\ref{tab:vbench}).
Our combined recipe recovers roughly half of that gap on subject and imaging metrics and surpasses FP16 on motion smoothness ($+$0.4\%), indicating that structured mixed precision plus smoothing is more effective than uniform 4-bit rounding for video diffusion FFNs.
Compared with using SmoothQuant alone, adding MixQ-style splitting specifically addresses the residual column spikes visible in Fig.~\ref{fig:wan-ffn} (bottom-left) that survive global folding.
Compared with MixQ-style splitting alone, SmoothQuant reduces the number of columns routed through the high-precision branch. This branch is narrow and deterministic; it is not equivalent to leaving whole Transformer blocks in FP16, but it still makes high-precision use explicit and easy to audit under the challenge constraints.

Remaining gaps to FP16 concentrate on aesthetic and subject-consistency scores, which are sensitive to fine appearance details that still clip under 4-bit activations even after fold.
We observe the largest activation tails in early high-noise FFN blocks; deeper layers (Fig.~\ref{fig:wan-ffn-deep}) show similar structure with slightly lower maxima, suggesting layer-wise $\alpha$ and $k$ could further close the gap without exceeding the organizer high-precision budget.
Future work could port token-local outlier prediction from full MixQ~\cite{mixq} to spatio-temporal diffusion tokens, and explore MXFP4 block shapes under the five-layer FP16 allowance.

\section{Reproducibility}

Code, calibration scripts, and evaluation commands are at \url{https://github.com/shch-y/icme}; \texttt{Install.md} documents environment setup and VBench reproduction.

\section{Conclusion}

We described a Wan2.2 I2V W4A4 inference stack that combines SmoothQuant-style folding, MixQ-style outlier splitting, and block-wise HiF4 packing.
Calibration on OpenS2V-5M shows heavy FFN activation tails; smoothing plus a narrow high-precision branch mitigates them enough to beat a native HiFloat4 baseline on all reported VBench I2V metrics while approaching FP16 quality.
The key lesson is to combine format-level 4-bit quantization with outlier-aware structure rather than treating W4A4 as uniform per-tensor rounding.

\section*{Acknowledgment}

We thank the organizers of the ICME 2026 Low-Bit-width Large-Model Quantization Challenge.
Our submission builds on Wan2.2, VBench, HiFloat4 tooling, SmoothQuant, MixQ, and OpenS2V-5M resources.

\bibliographystyle{IEEEbib}
\bibliography{icme2026references}

\end{document}